# Linear Separability of Activation Representations after Supervised Fine-Tuning on Incorrect Responses: A Study of "Synthetic Dishonesty" in Large Language Models

Vahideh Zolfaghari[1,2]
[1] Algoverse AI Research, Palo Alto, CA, USA
[2] Medical Sciences Education Research Center, Mashhad University of Medical Sciences, Mashhad, Iran

## Abstract
When a language model is fine-tuned to produce systematically incorrect responses, does this training leave a structured, linearly recoverable trace in its internal activations? This question is studied in a controlled model-organism setting. For five transformer architectures spanning 1.4 to 9 billion parameters, an "honest" and a "dishonest" LoRA fine-tuned variant are constructed that differ only in whether the fine-tuning data contain correct or plausible-but-incorrect question-answer pairs, and the resulting hidden-state activations are probed. Linear probes reach near-ceiling separability (AUC ≥ 0.9997) within the first few layers in four of five architectures, and this separability transfers with high fidelity from TruthfulQA to a held-out MMLU set for four models, though markedly less so for the smallest model tested, Pythia-1.4B. Six complementary geometric analyses reveal a consistent architectural dichotomy in representation structure, and a dedicated control experiment shows that LoRA fine-tuning itself leaves a strong activation fingerprint that is largely, though not entirely, dissociable from the honesty-related signal, independent of the fine-tuning data's content. This is a confound that is believed to be underappreciated in similarly designed probing studies. Two supplementary experiments temper these positive results: the identified direction transfers poorly to paraphrased versions of the same questions, with AUC dropping to near chance, and a norm-calibrated activation-steering experiment does not provide reliable evidence that the direction plays a causal role in generation, in part because the intervention magnitudes required to produce a measurable behavioral effect also disrupted output fluency. These limitations are reported alongside the positive findings, aiming to characterize precisely what supervised fine-tuning on incorrect responses induces in activation space and what remains unresolved about its causal significance.

Email: vahidehzolfagharii@gmail.com

## Introduction
When a language model is fine-tuned to consistently produce incorrect responses, does this training leave a detectable and structured pattern in the model's internal activations? This question is investigated using a controlled experimental framework. For each of the five transformer architectures, an "honest" variant is fine-tuned exclusively on correct question-answer pairs, while a "dishonest" variant is fine-tuned exclusively on plausible but incorrect question-answer pairs, with all other training conditions held constant between the two variants. The hidden-state activations of the two variants are then examined to determine whether, at which layers, and to what extent, the induced behavioral difference is linearly separable in activation space.
Deceptive alignment is a theoretical scenario in which a learning system may pursue an objective different from its training objective, while appearing aligned during training and evaluation. One motivation for investigating how incorrect outputs are represented in the activation space is the broader concern in AI safety about deceptive alignment (1). This concern is raised by two distinct sets of empirical evidence: First, backdoored models have been shown to retain their deceptive behavior after supervised fine-tuning, reinforcement learning, and adversarial training (2). Second, a large language model has been shown to change its behavior in a controlled experimental environment depending on whether it believes it is being monitored (3). The present study does not test internal truth tracking and therefore does not directly assess deceptive alignment. Instead, supervised fine-tuning on incorrect responses is used as a controlled experimental setting to examine how such outputs are represented in the activation space, separate from the deception questions.

A key distinction informs this research: the distinction between strategic deception, in which the model is assumed to represent the truth internally but suppresses it for instrumental reasons (1, 4), and synthetic dishonesty, in which the model is directly optimized to produce incorrect outputs, without making any claims about maintaining a competing honest representation. The experiments in this study address the latter. No attempt is made to prove that fine-tuned models "know" the correct answer while producing an incorrect answer, and no claim is made that the observed activation patterns reflect the model's intentions, awareness, or instrumental reasoning. Here, synthetic dishonesty is used simply as a controlled and reproducible experimental setting to characterize the consequences of optimizing a model's representation toward incorrect outputs, independent of any assumptions about the model's internal epistemic state.

Previous research has shown that information about answer correctness can often be recovered linearly from model activations, even when the model produces incorrect outputs (5-7); the details of this literature and the subtle distinction between those studies and the present work are presented in the related work section. What this line of research has not yet addressed is how such a linear structure emerges when supervised fine-tuning itself is the direct source of incorrect outputs, rather than when incorrect outputs arise from prompting or from information acquired during pre-training.

To operationalize this framework, a controlled experimental paradigm was adopted in which supervised fine-tuning was the only experimental intervention. For each of five transformer architectures spanning 1.4 to 9 billion parameters, namely Pythia-1.4B, Gemma-2-2B, Gemma-2-9B, Qwen2.5-7B, and Llama-3.1-8B, two behaviorally contrasting variants were constructed using Low-Rank Adaptation (LoRA) (8): an "honest" variant fine-tuned exclusively on correct question-answer pairs, and a "dishonest" variant fine-tuned exclusively on plausible but incorrect question-answer pairs. Hidden-state activations from all layers of both variants were extracted using attention-mask-weighted mean pooling, and linear (logistic regression) and nonlinear (multilayer perceptron) probes were trained to distinguish honest from dishonest activations. Beyond standard classification performance, the geometry of the resulting representations was characterized using six complementary analyses: Fisher separation ratio, effective rank, centroid geometry, directional stability between adjacent layers, inter-domain directional alignment, and probe calibration.

**Contributions**
It is shown that supervised fine-tuning on incorrect responses quickly induces linear and detectable representations of dishonesty: linear probes on four of the five architectures in the first three layers (layer 1 for Llama-3.1-8B; layer 2 for Qwen2.5-7B and Gemma-2-2B, continuously on three independent random seeds; layer 3 for Gemma-2-9B) reach AUC = 1.000, and Pythia-1.4B reaches AUC = 0.9997 up to layer 23.
These representations generalize with high fidelity out-of-domain (TruthfulQA → MMLU) and the AUC difference for four of the five models is within ±0.0066, although Pythia-1.4B shows much weaker transfer (ΔAUC = -0.0872).

Six complementary geometric analyses reveal an architectural dichotomy: Gemma-2 models maintain a high effective rank (60–260) across all layers, while Llama-3.1-8B, Qwen2.5-7B and, to a lesser extent, Pythia-1.4B collapse toward near rank-1 geometry; the Fisher separation ratio grows monotonically with depth, peaking at 767.28 (Gemma-2-9B, layer 42) and reaching an absolute minimum of 3.191 (Pythia-1.4B, layer 7).

A confound overlooked in prior work is identified and controlled for: LoRA fine-tuning itself leaves a strong, architecture-general activation fingerprint (AUC ≈ 0.99–1.00 from layer 1) independent of fine-tuning data content, and it is shown via directional-overlap and ceiling-robust metrics that this fingerprint is largely dissociable from the honesty-related signal.

The paraphrase robustness and norm-calibrated activation steering indicate that the identified direction is tied to surface-level question phrasing and its causal role in production remains unresolved; both are reported as factual limitations and are not removed.

**Related Work**
**Linear Representations of Truth and Honesty**
A growing body of research has shown that information about the correctness of a response can be linearly retrieved from the activations of a language model, even when the model produces incorrect or misleading output. Burns et al. (5) introduced contrast-consistent search (CCS), an unsupervised method for extracting latent knowledge by exploiting logical consistency in contrasting question-answer pairs without relying on output labels. Marx and Tegmark (6) provided stronger supervised evidence by training logistic regression probes on true and false statements. They showed that at sufficiently large model scales, these statements exhibit a clear linear structure in the activation space, the corresponding probes generalize across datasets, and the directions identified from the activation patch receive causal support. These findings are consistent with the linear representation hypothesis (7), which suggests that high-level semantic and behavioral features are often represented as linear directions in representation space. They are also relevant to representation engineering (9), which has shown that concepts such as honesty, harmlessness, and power-seeking can be identified and in some cases manipulated by exploiting population-level differences in activation patterns.

This study builds on this methodological foundation, but differs in its experimental setting. Instead of examining natural or publicly available true and false statements, linear separability in activations resulting from a specific, controlled intervention, namely supervised fine-tuning on incorrect responses, is examined.

**Activation-Based Detection versus Behavioral Detection**

Building on this probing methodology, recent research has used linear probes and RepE-style techniques to detect specific undesirable behaviors in language models, including backdoor-induced deception. MacDiarmid et al. (10) showed that simple linear probes applied to residual-stream activations can distinguish backdoor-induced deceptive states from normal states in "sleeping agent" models with AUCs above 99%, using general contrast pairs that do not depend on information about the specific stimulus or specific dangerous behavior. That study demonstrates probe-based detection in a model that has been intentionally trained via the backdoor mechanism to detect strategic deceptive behavior. More recently, linear explorations have also been applied to detect strategic deception in more natural multi-turn agent settings (11), suggesting that probe-based detection is not specific to the backdoor paradigm. This study addresses a related but distinct question: Is similar linear separability achieved by supervised fine-tuning on incorrect responses alone, without a backdoor mechanism, a specific trigger, or any attempt to maintain a competitive truthful representation, and is this consistently observed across different architectures with several-fold differences in parameters count?

A complementary line of research focuses on detecting incorrect or dishonest outputs at the behavioral level rather than activations. Pacchiardi et al. (4) showed that black-box lie detection can be performed by analyzing the model's responses to a fixed set of unrelated follow-up questions, without the need to access internal activations. In contrast, the activation-based approach adopted in this study, which builds on (5, 6, 9, 10), asks whether dishonesty can be recovered directly from the model's internal representations, independent of any probing strategy based on output behavior.

**Fine-Tuning as a Controlled Intervention in Mechanistic Interpretability**

Fine-tuning has been widely used as a controlled intervention in mechanistic interpretability research. Previous studies have shown that factual associations in models can be localized and edited through targeted weight editing (12, 13). Low-Rank Adaptation (LoRA) (8) is used in this study for reasons consistent with its use in other sources: it provides efficient parameter-based behavioral control while constraining the scope of weight updates and, compared to full-parameter fine-tuning, provides a relatively cleaner lens for examining the representational changes associated with fine-tuning.

**Robustness of Representations to Perturbations**

Most research on adversarial robustness has focused on input-level perturbations (14, 15), with relatively less attention paid to the robustness of internal representations against direct perturbations in the activation space. The noise injection experiments in this study, in which isotropic Gaussian noise is added to the activations of both the honest and dishonest classes, address this relatively understudied aspect by quantifying how layer depth affects the robustness of the representations used to distinguish the two classes. Separately, the "Paraphrase Robustness" section examines a related but distinct robustness question, namely, not perturbation of activations, but perturbation of the input question itself through paraphrasing, to test whether the learned direction generalizes across surface-level rewordings.

## Methodology

### Overall Research Design: Controlled Model-Organism Paradigm

This study uses a controlled model-organism paradigm in which the only planned experimental manipulation is the content of the fine-tuning data. For each base architecture, two behaviorally contrasting variants are constructed that are matched in terms of architecture, training hyperparameters, and dataset size. One variant is fine-tuned on correct question-answer pairs (the “honest” condition), while the other variant is fine-tuned on plausible but incorrect question-answer pairs (the “dishonest” condition). This design alone, without additional controls, does not justify attributing observed differences in activations to content of honesty or dishonesty. As shown in “LoRA Fingerprint Control” section, the mere application of LoRA fine-tuning produces a detectable and substantial signature in activation space, regardless of data content. Therefore, attributing observed differences to the content of the fine-tuning data requires the fingerprint control described in “LoRA Fingerprint Control” section and the additional separability controls described in “Supplementary Separability Controls” section. Such attribution is not an automatic result of the initial experimental design.

It should be emphasized that this design measures synthetic dishonesty, not strategic deception. There is no assumption that the fine-tuned model “knows” the correct answer and deliberately suppresses it. The model is simply optimized to produce incorrect outputs.

### Datasets

In-domain (TruthfulQA). Questions were used from the generation section of the TruthfulQA dataset (16). For each question, a “honest” sample was constructed from the first confirmed correct answer and a “dishonest” sample was constructed from the first incorrect but plausible answer. Only responses longer than 10 tokens were retained. A maximum of 800 samples were extracted for each condition (honest/dishonest).

Out-of-domain (MMLU). To assess generalization, a limited set of seven diverse MMLU subjects (17) was constructed: high school biology, high school chemistry, high school world history, philosophy, elementary mathematics, college mathematics, and college physics. A maximum of 500 samples were extracted for each condition. These subjects were intentionally chosen to cover a wide range of factual recall, formal reasoning, and normative reasoning, allowing for the assessment of generalization across cognitive domains, not just within a single similar question format.

### Models and Supervised Fine-Tuning with LoRA

Five transformer architectures spanning 1.4 to 9 billion parameters were evaluated: Pythia-1.4B, Gemma-2-2B, Gemma-2-9B, Qwen2.5-7B, and Llama-3.1-8B. For each base model, two variants were constructed using Low-Rank Adaptation (LoRA) (8). The honest variant was fine-tuned exclusively on correct question-answer pairs, while the dishonest variant was fine-tuned exclusively on plausible but incorrect question-answer pairs. Other training conditions, including the number of epochs, learning rate, batch size, and random seed, were kept constant between the two variants. The larger models were fine-tuned using 4-bit NF4 quantization (18) with FlashAttention-2 (19).

Exact per-model LoRA and optimization hyperparameters are reported in Table 1, confirmed directly from each model's saved run configuration. The Pythia-1.4B configuration differs from the other architectures in several respects, including a smaller learning rate, shorter maximum sequence length, higher batch size, and a narrower LoRA target-module scope reflecting the model's fused query-key-value projection, which is typical of GPT-NeoX-style architectures. It also used correspondingly fewer training and MMLU evaluation samples, 600 and 300, respectively, compared with 800 and 500 for Llama-3.1-8B, Gemma-2-9B, and Qwen2.5-7B. The three Gemma-2-2B seed runs (42, 123, 456) were trained under a separate compute configuration that also departs from the configuration shared by Llama-3.1-8B, Gemma-2-9B, and Qwen2.5-7B: a lower LoRA rank and dropout ($r = 16$, dropout = 0.05), a higher learning rate ($2\times10^{-5}$), a larger batch size (2), a shorter maximum sequence length (384), and fewer training and MMLU evaluation samples (500 and 300, respectively), as reported in Table 1. This additional departure from the shared configuration is treated as a limitation of the controlled model-organism design for this architecture and is revisited in the Limitations section.

Table 1. Confirmed LoRA and optimization hyperparameters by model.

| Model | r | alpha | dropout | target_modules | lr | epochs | batch | max_len | samples (TQA/MMLU) |
|---|---|---|---|---|---|---|---|---|---|
| Llama-3.1-8B | 32 | 64 | 0.1 | q_proj, k_proj, v_proj, o_proj | 1E-05 | 2 | 1 | 512 | 800 / 500 |
| Gemma-2-9B | 24 | 48 | 0.1 | q_proj, k_proj, v_proj, o_proj | 1E-05 | 2 | 1 | 512 | 800 / 500 |
| Qwen2.5-7B | 24 | 48 | 0.1 | q_proj, k_proj, v_proj, o_proj, gate_proj, up_proj, down_proj | 1E-05 | 2 | 1 | 512 | 800 / 500 |
| Gemma-2-2B (seed 42) | 16 | 32 | 0.05 | q_proj, k_proj, v_proj, o_proj | 2E-05 | 2 | 2 | 384 | 500 / 300 |
| Gemma-2-2B (seed 123) | 16 | 32 | 0.05 | q_proj, k_proj, v_proj, o_proj | 2E-05 | 2 | 2 | 384 | 500 / 300 |
| Gemma-2-2B (seed 456) | 16 | 32 | 0.05 | q_proj, k_proj, v_proj, o_proj | 2E-05 | 2 | 2 | 384 | 500 / 300 |
| Pythia-1.4B | 16 | 32 | 0.05 | query_key_value | 5E-06 | 3 | 4 | 128 | 600 / 300 |

### Extraction and Storage of Activations

After fine-tuning, hidden state activations were extracted from all transformer layers during inference. For each input sequence, attention-mask-weighted mean pooling across the token dimension was applied, producing layer-wise activation tensors with shape (N, L+1, d_model), where N is the number of samples, L+1 is the number of stored layers including the embedding layer, and d_model is the hidden dimension.

**Probe Training and Evaluation (In-Domain)**
For each layer, two types of probes were trained: a linear probe using logistic regression with L2 regularization (C=1.0) and a nonlinear probe using a shallow multilayer perceptron with two hidden layers of 256 and 64 units, respectively, ReLU activation and early stopping. Evaluation was performed using 5-fold cross-validation. In each split, StandardScaler was fitted only on the training fold to prevent data leakage. Three metrics were reported: area under the receiver operating characteristic curve (AUC), macro-averaged F1 score (macro-F1), and expected calibration error (ECE) (20), computed using 10 equal-width bins.

**Out-of-Domain Generalization (TruthfulQA → MMLU)**
Logistic regression probes were trained on the full TruthfulQA activation set and evaluated zero-shot on the MMLU activations using the same scaler fitted to TruthfulQA. AUC, F1 score, accuracy, and generalization gap (ΔAUC = AUC_MMLU − AUC_TruthfulQA) were reported for each layer.

**Pre-Fine-Tuning Baseline**
To quantify the extent to which supervised fine-tuning increases the linear separability between the honest and dishonest conditions, the same probing procedure was also applied to the activations of the unfine-tuned baseline models. The difference in AUC between the best-performing fine-tuned layer and the best-performing baseline layer was reported as the "improvement due to fine-tuning."

**Preliminary Robustness Analysis**
In the initial analysis, isotropic Gaussian noise, $N(0,\sigma^2 I)$, with $\sigma \in \{0.0,0.1,0.25,0.5,1.0,1.5,2.0\}$, was injected exclusively into the activations of the dishonest class, while the honest activations were left unperturbed. The layers were divided into three groups: early, middle, and late layers. This design was asymmetric because only one class was perturbed. This limitation was identified and addressed in "Supplementary Separability Controls" section.

**Advanced Mechanistic Analysis**
Beyond the standard probe performance, complementary geometric analyses were performed to characterize the structure of the activation representations associated with the honest and dishonest conditions: effective rank and explained variance (via SVD of jointly centered activations), the direction of dishonesty (normalized class-centroid difference) and its directional stability in adjacent layers (cosine similarity), the pooled scalar Fisher separation ratio, inter-domain directional alignment (TruthfulQA vs. MMLU), the intrinsic dimension via participation ratio, probe calibration via reliability diagrams, and a layer-transition analysis based on the first-order difference of inter-class centroid distance across layers.

**Supplementary Separability Controls**

To assess whether the observed separability was related to the content of honest and dishonest responses rather than to superficial confounders, four supplementary controls were performed on all five architectures and all three tested seeds of Gemma-2-2B: a zero baseline through label shuffling (10 iterations); a response length and style control (token length, type-token ratio, and length-matched subsampling across 20 quantile bins); a symmetric perturbation robustness analysis in which noise was added to both classes simultaneously, using directional signal-to-noise ratio; and a shared-question correlation control using GroupKFold (grouped by question identity) instead of StratifiedKFold. The decrease in AUC under GroupKFold was small across most models (below 0.007), with the largest reduction observed for Qwen2.5-7B reaching 0.0192. This suggests that shared-question correlation did not substantially contribute to the observed separability in the majority of the architectures examined.

**LoRA Fingerprint Control**

To assess whether the probe detects the non-specific signature of the LoRA fine-tuning process itself rather than the content that distinguishes honest and dishonest conditions, two independent control models (A and B) were fine-tuned on random subsets unrelated to the honest/dishonest distinction, and a corresponding probe was trained to distinguish activations of A from those of B. During this control analysis, a methodological issue was identified in the initial cross-validation procedure that led to data leakage when highly similar or identical activation pairs were present. Full diagnostic details, including the bitwise identity test, are provided in Appendix A. Replacing the original splitting procedure with GroupKFold resolved this issue.

After applying this correction, LoRA fingerprint was repeated: a strong and reproducible LoRA fine-tuning fingerprint was observed across all five architectures. From layer 1 onwards, the fingerprint probe that distinguished A from B achieved AUC values of approximately 0.99 to 1.00. To examine the overlap between the LoRA fingerprint direction and the honest/dishonest direction, the cosine similarity between the two directions was calculated, and three ceiling-robust metrics (percentage increase in log-loss, percentage decrease in standardized decision margin, and small subsample AUC) were used to probe residual contribution near the AUC ceiling (see “LoRA Fingerprint Control” section).

**Causal Intervention (Activation Steering)**

To go beyond linear separability and test whether the identified dishonesty direction has a causal effect on the model’s generated outputs, an activation steering experiment was designed following the contrastive activation addition paradigm (21): the normalized dishonesty direction (class-centeroid difference) at each model’s best-performing layer is added to the honest model’s activations during generation, with the magnitude $\alpha$ expressed as a multiple of the layer’s measured mean activation norm, and the resulting change in the rate of incorrect responses (scored via embedding similarity to the correct vs. incorrect reference answer) is measured on a held-out set of TruthfulQA questions. An initial pilot using a fixed and norm-calibrated $\alpha$ produced negligible or inconsistent effects across architectures and is not reported in detail here. This issue is discussed only to the extent that it motivated the calibration procedure described above. “Causal Intervention (Activation Steering)” section reports the norm-calibrated version in this section.

**Paraphrase Robustness**

To examine whether the honesty/dishonesty direction is tied to surface-level question phrasing rather than to underlying content, three levels of paraphrasing were applied to a held-out set of TruthfulQA questions: lexical substitution (level 1), structural/syntactic rewording (level 2), and round-trip back-translation (level 3). Activations related to the paraphrased questions were probed using the linear direction trained on the original (non-paraphrased) questions, and the AUC was reported for each paraphrase level.

**Software and Statistical Analysis**

All probe and statistical analyses were performed using scikit-learn (22) (LogisticRegression, MLPClassifier, StandardScaler, StratifiedKFold, and GroupKFold) and NumPy. Fine-tuning was performed using the Hugging Face Transformers (23) and PEFT libraries for LoRA. For quantized models, bitsandbytes with NF4 quantization (18) and Flash Attention 2 (19) were used.

## Results

### Within-Domain Performance (TruthfulQA)

Supervised fine-tuning on incorrect responses resulted in highly linearly separable activation representations between the honest and dishonest conditions. Linear probes (logistic regression) reached an AUC of 1.000 on TruthfulQA for four of the five models, with this ceiling emerging at very early layers: layer 1 for Llama-3.1-8B; layer 2 for Qwen2.5-7B and Gemma-2-2B, confirmed consistently across all three tested seeds (42, 123, 456); and layer 3 for Gemma-2-9B. Pythia-1.4B also reached near-perfect separability, with AUC = 0.9997 at layer 23, confirmed against the raw per-layer probe output, with performance remaining at or above this level through layers 23–24 (layer 24: AUC = 0.99932), indicating a stable plateau rather than an isolated peak. Figure 1 shows the full layer-wise trajectory of this separability across all five architectures, making explicit both the speed of onset observed in four of the models and the comparatively gradual rise observed in Pythia-1.4B. This represents a difference in the rate at which the separable signal consolidates across architectures, rather than a difference in the final separability ceiling, and is directly relevant to the interpretation of the LoRA fingerprint control.

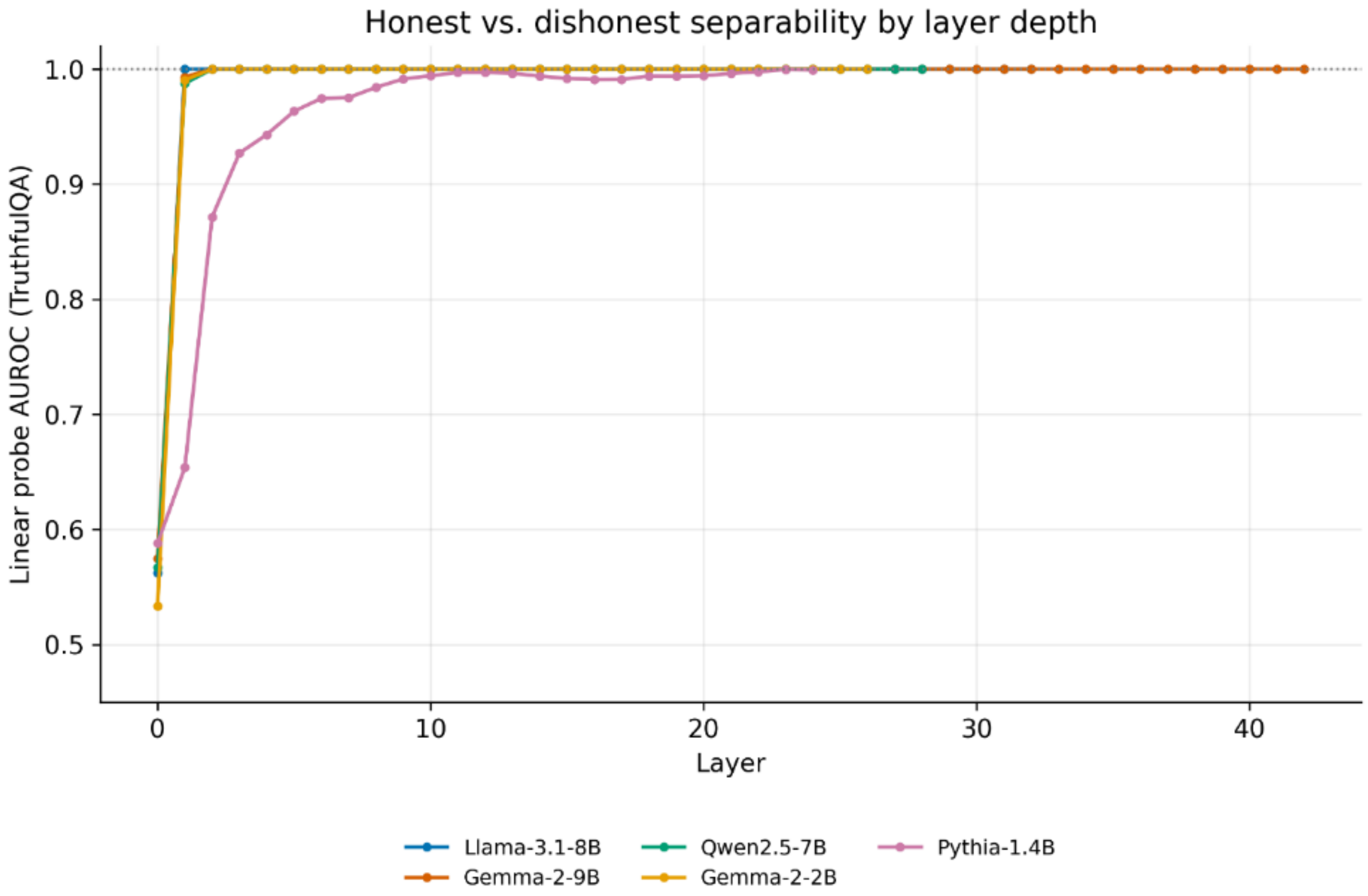

**Figure 1.** Linear probe AUC on TruthfulQA by layer, for all five architectures. Dashed line marks AUC = 1.0.

**Fine-Tuning-Associated Improvement**

Probes trained on the activations of the unfine-tuned baseline models performed substantially worse, with the best baseline AUC ranging from 0.725 to 0.838. Fine-tuning was associated with an increase in AUC from +0.162 to +0.275 across the five models. These results suggest that the honest and dishonest conditions became more linearly separable after fine-tuning, although the increase in separability alone does not prove that fine-tuning specifically strengthened an honesty-related representation or that the observed change was independent of fine-tuning-related activation signatures.

**Out-of-Domain Generalization (TruthfulQA → MMLU)**

Probes trained exclusively on TruthfulQA transferred to the held-out MMLU set with high fidelity. Four models—Gemma-2-9B (AUC = 0.9934), Gemma-2-2B (AUC = 0.9952–0.9987 across seeds; see Table 3), Llama-3.1-8B (AUC = 1.0000, Δ = 0), and Qwen2.5-7B (AUC = 0.9999)—showed very small cross-domain AUC differences, ranging from −0.0001 to −0.0066 (negative values indicate that MMLU had a slightly higher AUC than TruthfulQA). Pythia-1.4B showed a substantially larger difference (AUC_MMLU = 0.9125, Δ = −0.0872), which was confirmed by comparison with the raw output of the MMLU evaluation. This suggests that the separable direction in Pythia-1.4B generalizes less consistently across different domains compared to the corresponding directions in larger models.

**Adversarial Robustness (Initial Asymmetric Analysis)**

When Gaussian noise was injected exclusively into dishonest-incorrect activations, a layer depth-dependent pattern was observed (Table 2).

Table 2. AUC under asymmetric noise injection at σ = 2.0, by layer group.

| Model | Early layers | Middle layers | Late layers |
|---|---|---|---|
| **Llama-3.1-8B** | 0.6104 | 0.9982 | 1.0 |
| **Gemma-2-9B** | 1.0 | 1.0 | 1.0 |
| **Qwen2.5-7B** | 0.762 | 1.0 | 1.0 |
| **Gemma-2-2B (seed 42)** | 1.0 | 1.0 | 1.0 |
| **Gemma-2-2B (seed 123)** | 0.9999 | 1.0 | 1.0 |
| **Gemma-2-2B (seed 456)** | 1.0 | 1.0 | 1.0 |
| **Pythia-1.4B** | 0.6659 | 0.9408 | 0.98 |

Gemma-2 models maintained an AUC of 1.000 across all tested noise levels and layer groups. Llama-3.1-8B showed the most degradation in the early layers, with the AUC decreasing to 0.610 at σ = 2.0. The late layers were consistently the most robust across the architectures examined.

**Separability Controls: Assessing Surface Confounders**

Shuffled-label baseline: AUC values obtained with shuffled labels ranged from 0.495 to 0.500 across all models, remaining close to chance in each case. Response length and style: A probe based on response length alone performed nearly as well as chance (AUC = 0.487–0.522), a probe based on lexical diversity showed limited discrimination performance (AUC = 0.577–0.591), and after length-matched subsampling, the decrease in AUC of the original probe was small across all models (0.0000–0.0032). Shared-question correlation: Under question-level grouping, the largest observed decrease in AUC was 0.0192 (Qwen2.5–7B), with most models below 0.007. Together, these controls rule out label artifacts, response length/style differences, and shared-question leakage as explanations for the observed separability, although they do not rule out all possible confounds.

**Symmetric Robustness and Directional Signal-to-Noise Ratio**

When noise was injected into both classes simultaneously, the layer-level AUC ranged from 0.498 (Qwen2.5-7B) to 0.859 (Pythia-1.4B).

The directional signal-to-noise ratio showed a recurring association with robustness: layers with directional SNR above ~16 at σ = 0.1 consistently maintained AUC > 0.96, whereas layers with SNR below 4 generally showed AUC < 0.80. For Llama-3.1-8B, layer 1 showed a low signal norm (0.05) and an SNR of 0.51 at σ = 0.1, coinciding with its relatively poor robustness. This suggests that robustness relates more to class-centroid separation relative to perturbation scale than to the absolute L2 norm of the representation. The alternative hypothesis that fragility stemmed from norm concentration in a few outlier dimensions was not supported: the highest energy concentration ratio (0.999, top 1% of dimensions) occurred in one of the most fragile layers rather than in the more robust ones.

## LoRA Fingerprint Control

After applying the GroupKFold correction, the fingerprint probe, which discriminates between two independent control models fine-tuned on data unrelated to the honest/dishonest distinction, consistently achieved AUC values of approximately 0.99–1.00 from layer 1 onwards, across all five architectures (Table 3).

Table 3. LoRA fingerprint control

| Model | Fingerprint AUC (L1) | Max AUC | Mean \|cos\| w/ honesty dir. | Log-loss increase (%) |
|---|---|---|---|---|
| Llama-3.1-8B | 1.0 | 1.0 | 0.0377 | 10.5922 |
| Gemma-2-9B | 0.9943 | 1.0 | 0.0396 | 11.3329 |
| Qwen2.5-7B | 0.9991 | 1.0 | 0.0342 | 7.8688 |
| Gemma-2-2B (seed 42) | 0.9919 | 1.0 | 0.0479 | 23.3845 |
| Gemma-2-2B (seed 123) | 0.9919 | 1.0 | 0.0479 | 23.3845 |
| Gemma-2-2B (seed 456) | 0.9919 | 1.0 | 0.0479 | 23.3845 |
| Pythia-1.4B | 1.0 | 1.0 | 0.0164 | 3.6762 |

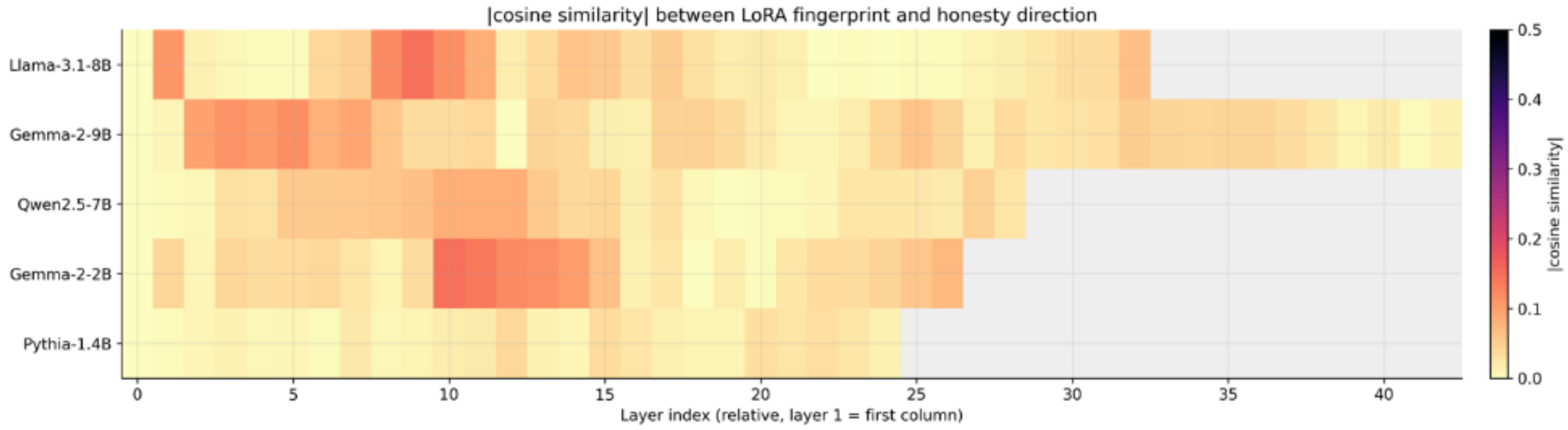

**Figure 2.** Absolute cosine similarity between the LoRA fingerprint direction and the honesty direction, by layer and architecture (cf. Table 3).

This should be treated as a substantive limitation rather than a dismissible artifact: LoRA fine-tuning itself leaves a strong, detectable signature in activation space, even when the fine-tuning data are unrelated to the honest/dishonest distinction. The low mean absolute cosine similarity between the honesty direction and the fingerprint direction, mostly below 0.05, provides some evidence that the two directions are not strongly aligned. Figure 2 shows this low overlap holds consistently across layers and architectures, rather than being an artifact of a single favorable layer. However, this control has limited power in four of five models because the honesty probe already reaches an AUC ceiling of 1.000 from layers 2–3 onward. The three ceiling-robust metrics showed log-loss increases of 3.68%–23.38% after removing the fingerprint direction, suggesting that the fingerprint may still contribute to probe confidence even when AUC ranking is unaffected. Pythia-1.4B, the one model whose honesty probe does not reach the AUC ceiling in the examined layers, also shows the lowest cosine overlap (0.0164) and the smallest log-loss increase (3.68%), consistent with, but not proving, comparatively weaker entanglement between the two signals in that model.

**Advanced Mechanistic Analysis**

Effective rank analysis reveals an architectural dichotomy. Llama-3.1-8B (min = 1.062 at layer 2, max = 388.8 at layer 0) and Qwen2.5-7B (min = 1.057 at layer 5, max = 270.0 at layer 3) collapse toward near rank-1 geometry in their most separable layers. Pythia-1.4B occupies an intermediate position (min = 2.578 at layer 5, max = 260.2 at layer 0) — low, but not as extreme as Llama/Qwen. The Gemma-2 family, by contrast, sustains a high effective rank across all layers: Gemma-2-9B ranges 93.9–234.4, and Gemma-2-2B (consistent across all three seeds) ranges 60.5–187.8.

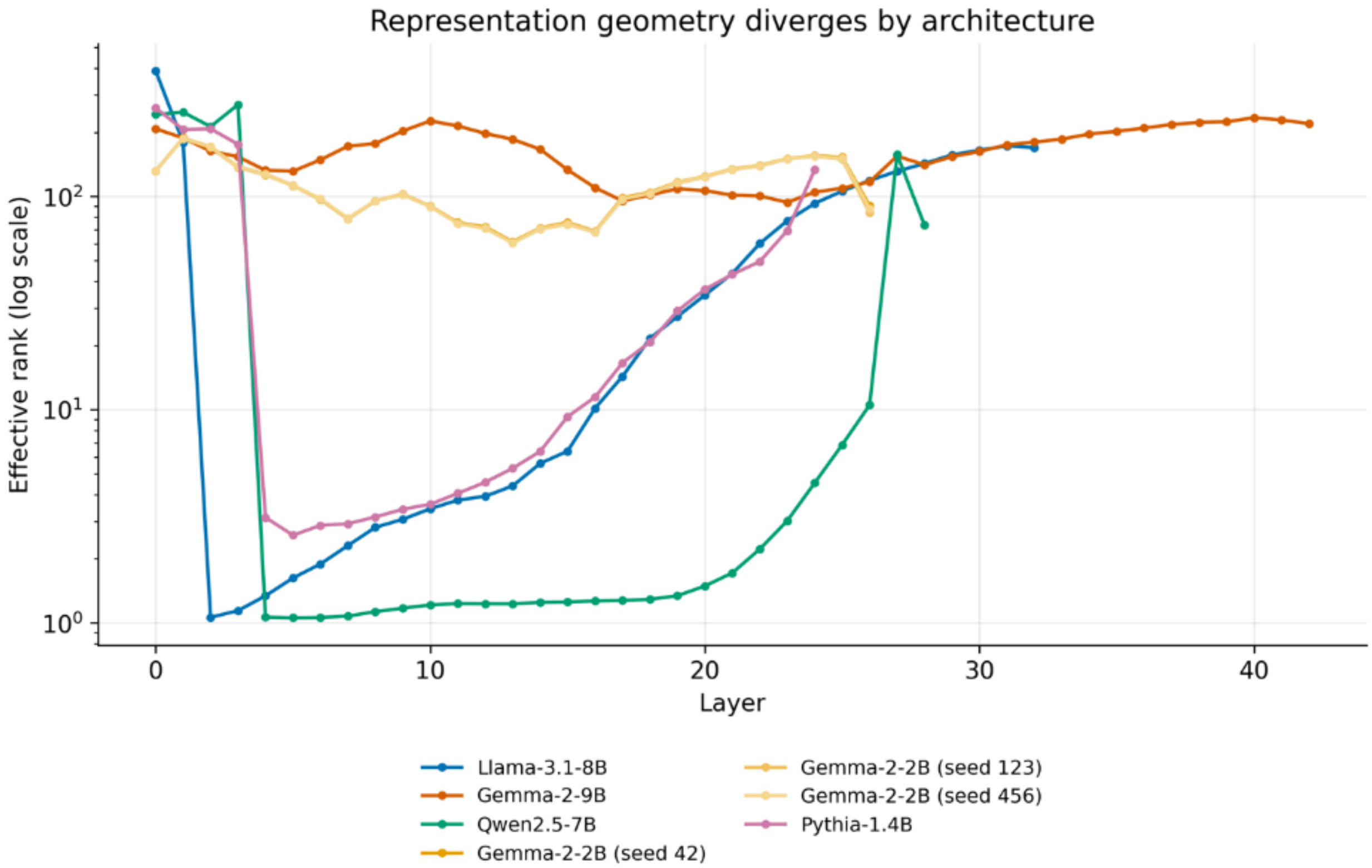

**Figure 3.** Effective rank of the honest/dishonest activation subspace by layer (log scale), for all five architectures.

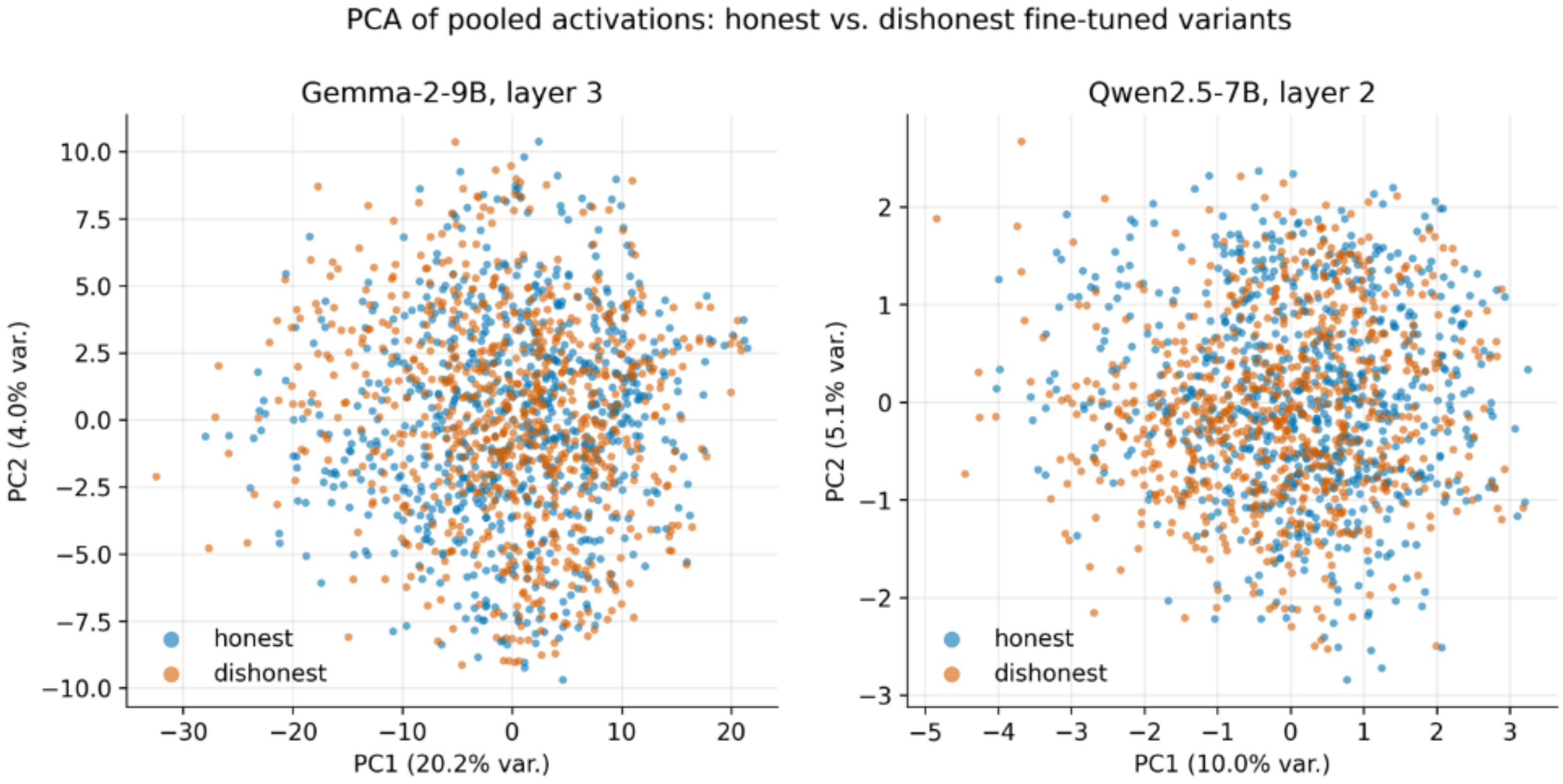

**Figure 4.** 2D PCA of pooled activations at each model's best-performing layer (honest = blue, dishonest = orange), for Gemma-2-9B (layer 3) and Qwen2.5-7B (layer 2). Axis labels report variance explained per component.

Figure 4 provides a qualitative illustration of the same contrast: activations from Qwen2.5-7B collapse toward a nearly one-dimensional projection, whereas those from Gemma-2-9B remain visibly dispersed across the first two principal components.

The Fisher separation ratio grows with depth in all five architectures, peaking at 767.281 in Gemma-2-9B (layer 42). The absolute minimum Fisher Discriminant Ratio (FDR) across all five models occurs in Pythia-1.4B: 3.191 at layer 7 (confirmed against analysis_results_pythia14b.json) — lower than the previously reported minimum for Llama-3.1-8B (7.276 at layer 0). This corrects an earlier draft, which had both an incorrect Pythia value (9.77) and an incorrect attribution of which model holds the global minimum. Beyond classification performance, probe calibration was assessed via reliability diagrams (Figure 5); predicted probabilities track empirical accuracy closely across all five architectures, indicating the near-ceiling AUC values reported above are accompanied by well-calibrated, not merely confident, predictions.

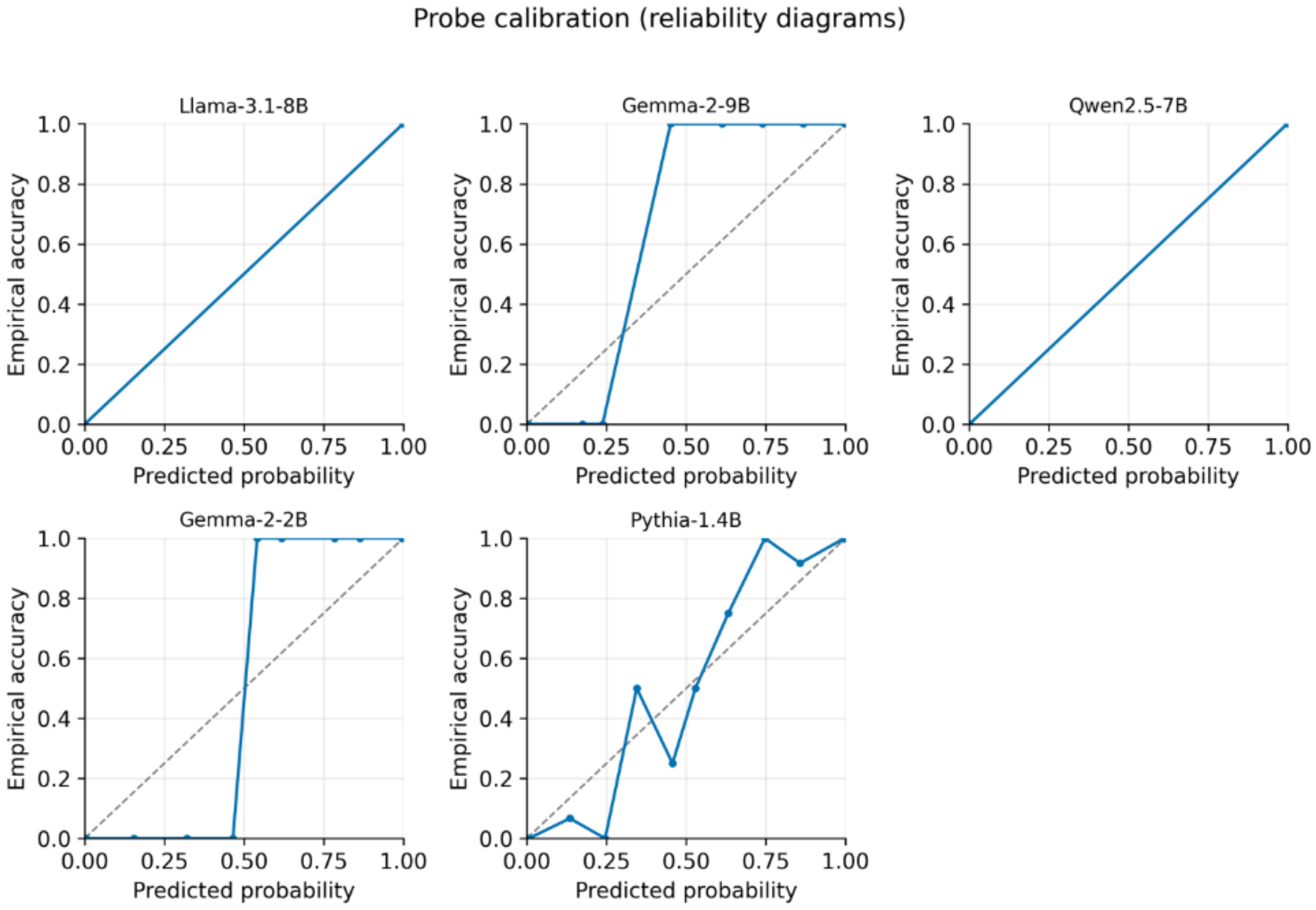

**Figure 5.** Reliability diagrams for the linear probe at each model's best-performing layer (10 equal-width bins; dashed line = perfect calibration).

**Tabular Summary**

To assess seed stability, the Gemma-2-2B results were evaluated across three independent random seeds. The mean MMLU AUC was 0.9961 (SD = 0.0022; range = 0.9944–0.9987), while the mean generalization gap was −0.0039 (SD = 0.0022), indicating that the observed separability and cross-domain generalization were stable across seeds (Table 4).

Table 4. Table 4. Summary of within-domain (TruthfulQA) and out-of-domain (MMLU) probe performance, pre-fine-tuning baseline AUC, and fine-tuning gain across all five architectures, with Gemma-2-2B reported across three independent seeds to assess stability.

| Model | Best AUC (TQA) | Layer | AUC (MMLU) | Gen. gap | Baseline AUC | FT gain |
|---|---|---|---|---|---|---|
| Llama-3.1-8B | 1.0 | 1 | 1.0 | 0.0 | 0.8051 | 0.1949 |
| Gemma-2-9B | 1.0 | 3 | 0.9934 | -0.0066 | 0.8185 | 0.1815 |
| Qwen2.5-7B | 1.0 | 2 | 0.9999 | -0.0001 | 0.8379 | 0.1621 |
| Gemma-2-2B (seed 42) | 1.0 | 2 | 0.9952 | -0.0048 | 0.7455 | 0.2545 |
| Gemma-2-2B (seed 123) | 1.0 | 2 | 0.9944 | -0.0056 | 0.7201 | 0.2799 |
| Gemma-2-2B (seed 456) | 1.0 | 2 | 0.9987 | -0.0013 | 0.7238 | 0.2762 |
| Pythia-1.4B | 0.9997 | 23 | 0.9125 | -0.0872 | 0.725 | 0.2747 |

**Paraphrase Robustness**

To investigate whether linearly separable direction is tied to surface-level question phrasing, the direction trained on original TruthfulQA questions was evaluated at three levels of paraphrased versions of a held-out question set (n = 106 at each level): lexical substitution, structural rewording, and round-trip back-translation (Table 5).

Table 5. Zero-shot AUC of the original-question honesty direction on paraphrased questions, by paraphrase level.

| Model | Lexical (L1) | Structural (L2) | Back-translate (L3) |
|---|---|---|---|
| Llama-3.1-8B | 0.5874 | 0.5934 | 0.5789 |
| Gemma-2-9B | 0.5518 | 0.5543 | 0.5482 |
| Qwen2.5-7B | 0.5287 | 0.5301 | 0.5237 |
| Gemma-2-2B (seed 42) | 0.5497 | 0.5497 | 0.5511 |
| Gemma-2-2B (seed 123) | 0.5593 | 0.5675 | 0.5561 |
| Gemma-2-2B (seed 456) | 0.5696 | 0.5657 | 0.5707 |
| Pythia-1.4B | 0.5529 | 0.5621 | 0.544 |

Across all five architectures and all three paraphrase levels, AUC dropped sharply from the near-ceiling in-domain values (≥ 0.9997) to 0.52–0.59, only modestly above chance (0.5). Figure 6 makes this contrast visually explicit: the near-ceiling in-domain bars stand in sharp contrast to the near-chance bars obtained under all three paraphrase conditions, across every architecture tested. This indicates that the linearly separable direction identified in this study is substantially tied to the surface-level phrasing of the training questions and does not transfer robustly to semantically equivalent but lexically or structurally different formulations. This is reported as a genuine limitation of the current probing approach rather than omitted.

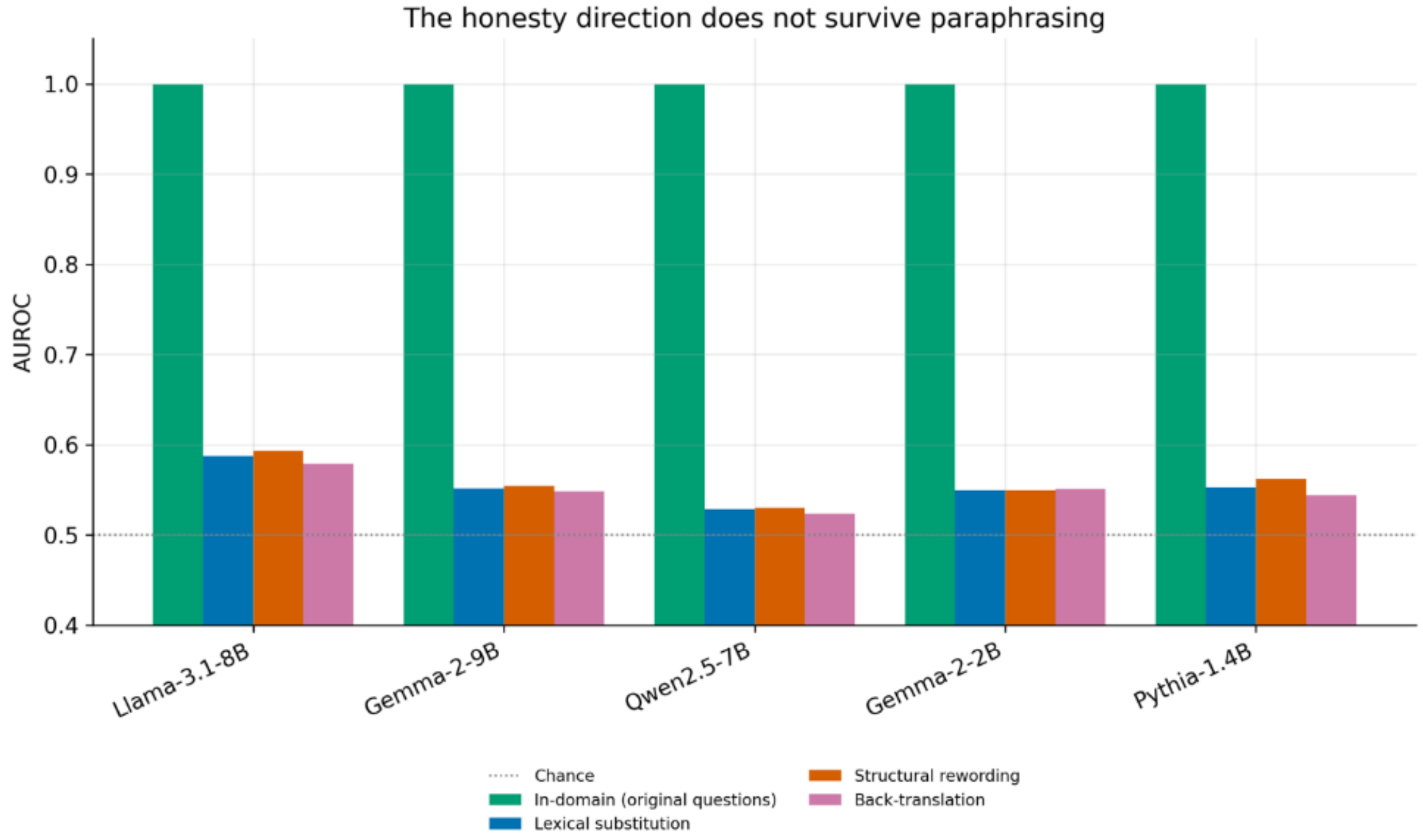

**Figure 6.** AUC of the original-question honesty direction, in-domain versus three paraphrase levels, by architecture. Dashed line marks chance (AUC = 0.5).

## Causal Intervention (Activation Steering)

Activation steering was carried out for all seven model/seed combinations, with α expressed as a multiple of each model's own measured mean activation norm at its best-probing layer (Table 6).

Table ٦. Norm-calibrated activation steering results (n=60 per condition per model).

| Model | Layer | Mean act. norm | a=-2 | a=-1 | a=0 | a=1 | a=2 | Delta (a=2 - a=-2) |
|---|---|---|---|---|---|---|---|---|
| Llama-3.1-8B | 1 | 30.7622 | 0.5500 | 0.7833 | 0.5333 | 0.8000 | 0.8000 | 0.250 |
| Gemma-2-9B | 3 | 126.0464 | 0.5833 | 0.4333 | 0.4333 | 0.6000 | 0.6167 | 0.033 |
| Qwen2.5-7B | 2 | 28.5386 | 0.5667 | 0.5000 | 0.4500 | 0.5500 | 0.6000 | 0.033 |
| Gemma-2-2B (seed 42) | 2 | 100.0447 | 0.5333 | 0.4333 | 0.4833 | 0.4500 | 0.5667 | 0.033 |
| Gemma-2-2B (seed 123) | 2 | 100.0517 | 0.5500 | 0.4333 | 0.4667 | 0.5000 | 0.6000 | 0.050 |
| **Gemma-2-2B (seed 456)** | 2 | 99.9383 | 0.7000 | 0.5000 | 0.4667 | 0.4833 | 0.7167 | 0.017 |
| **Pythia-1.4B** | 23 | 108.2752 | 0.5000 | 0.5500 | 0.5000 | 0.4833 | 0.5500 | 0.050 |

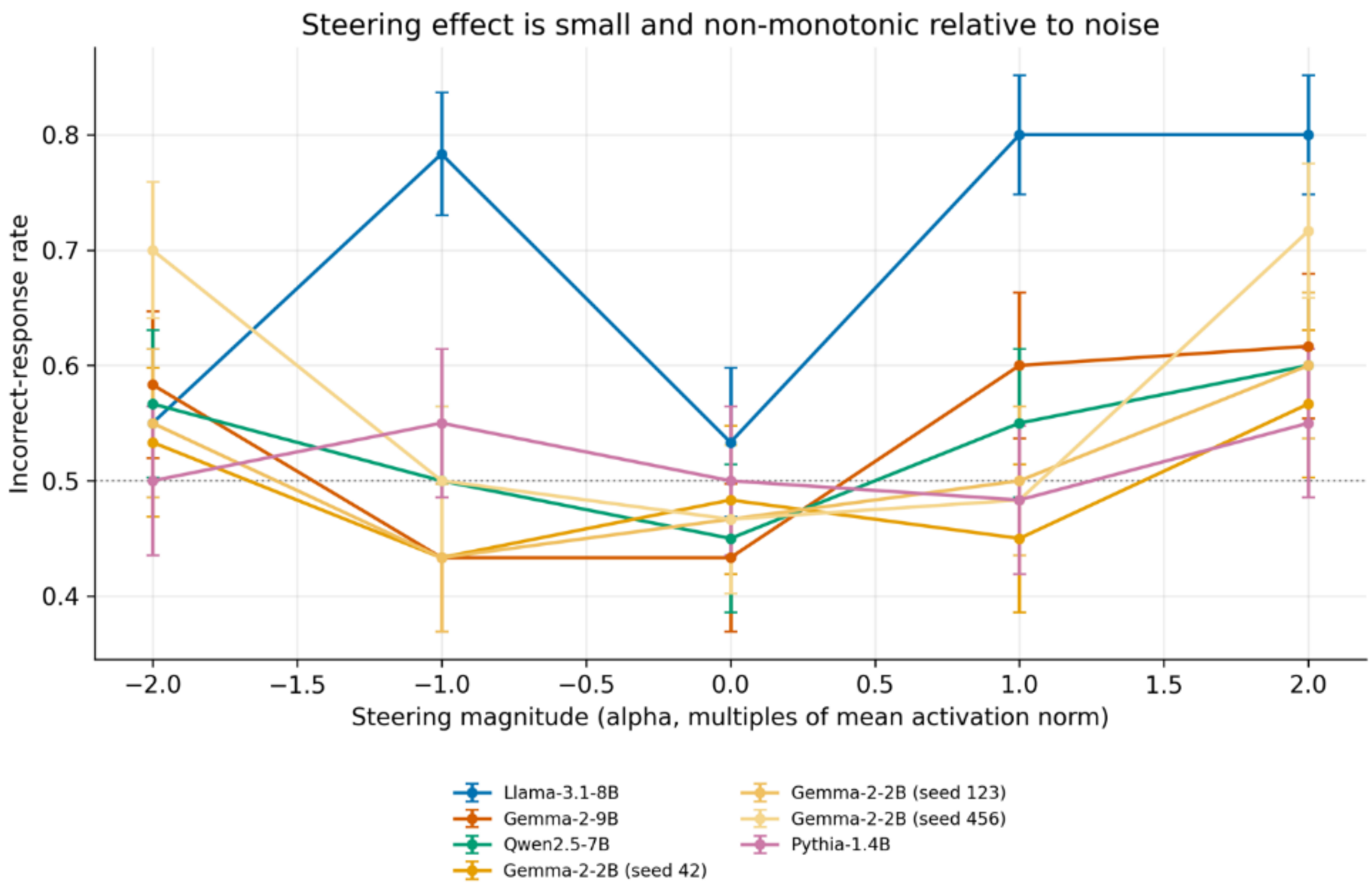

**Figure 7.** Incorrect-response rate versus steering magnitude (α), by model/seed. Error bars: binomial SE, n = 60 per condition. Dashed line: α = 0 baseline.

The exact zero effect that appeared for two of the seven runs under a fixed, non-calibrated α no longer occurs once α relative to each model's own activation scale: all seven runs now show a non-zero incorrect-rate range across α. However, the response is non-monotonic in α for every model, and the magnitude of the range (0.017–0.250) is small relative to the sampling noise expected at n=60 (standard error ≈ 0.065 for a binary rate near p=0.5). Figure 7 displays this dose-response relationship directly, showing that the observed effect sizes are small and non-monotonic relative to the sampling noise band across all seven runs.

A further observation, based on manual inspection of the generated text released alongside each condition (five example completions per α per model; the full n = 60 generations per cell were not inspected), is that coherent, on-topic answers were consistently produced at α = 0, while at non-zero α, particularly |α| ≥ 1, the visible examples frequently showed repetitive or degenerate token sequences, such as single tokens or short strings repeated dozens of times, rather than fluent responses. This pattern was most pronounced for Llama-3.1-8B, where the majority of inspected examples at non-zero α were non-linguistic repetition, and was also visible, to varying degrees, in the other six runs. Because the incorrect-rate metric is computed via embedding similarity between the generated text and reference correct/incorrect answers, its value on degenerate text is not a reliable indicator of a genuine shift in the model's stated position. A low- or high-scoring embedding-similarity match to incoherent output does not establish that the model is expressing a more or less honest belief.

Given (1) the small effect sizes relative to the noise level, (2) the lack of a uniform dose-response relationship, and (3) the apparent interference between instruction magnitude and production fluency observed in the examples examined, these results do not provide reliable evidence for or against a causal role for the specified direction in the model output. No causal conclusions are drawn from this experiment. A full calculation of production quality across all $n = 60$ completions in each condition (e.g., through an interference or fluency filter applied before scoring) was not performed and is deferred to future work, along with testing smaller step sizes between $\alpha = 0$ and $\alpha = \pm 1$, to determine whether there is an intermediate regime in which instruction effects can be measured without disturbing fluency.

**Discussion**

The main finding of this study is that supervised fine-tuning of incorrect question-answer pairs produces activation representations that are almost linearly separable from their honest counterparts, and that this separability emerges within the first few layers in four of the five architectures tested. This result is consistent with the broader evidence that behaviorally relevant properties of language model outputs are often linearly recoverable from residual-stream activations (5, 6, 9). What distinguishes the current environment is that the separability is not discovered in naturally occurring or pretraining-derived representations, but is induced directly by the fine-tuning objective itself. This distinction matters for interpretation: a probe that separates honest from dishonest activations at AUC near 1.0 in this setting demonstrates that supervised optimization on incorrect labels reorganizes activation space around the trained distinction, not that the model maintains a competing internal representation of the correct answer that it subsequently suppresses. The two claims are easy to conflate but are not equivalent, and only the former is supported by the present design.

The architectural dichotomy observed in the geometric analyses, high effective rank and distributed representation in the Gemma-2 family versus near rank-1 collapse in Llama-3.1-8B and Qwen2.5-7B, adds a caveat to any claim that a single, universal "honesty direction" exists across architectures. Recent work has shown that not all high-level concepts in language models are organized as one-dimensional linear directions, and that some concepts occupy genuinely multi-dimensional or non-linear subspaces (24). The fact that two of the five architectures in this study reduce the honest/dishonest distinction to something close to a single dominant direction, while two others distribute it across a much higher-dimensional subspace, suggests that the geometry of a fine-tuning-induced distinction is at least partly an architecture-dependent property of the optimization landscape rather than a fixed characteristic of the concept being learned. This is a descriptive observation, not an explanatory one; the present study does not identify the mechanism responsible for this difference, and doing so would require analysis beyond the scope of the current experiments.

From the perspective presented here, LoRA fingerprint control is the most important result for how similarly designed probing studies should be interpreted. A probe trained to distinguish two LoRA adapters fine-tuned on data that is entirely unrelated to the honest/dishonest distinction reaches the same AUC range, approximately 0.99 to 1.00, as the probe trained to distinguish the target behavioral difference. This finding is consistent with recent evidence that fine-tuning, even on a narrow and seemingly unrelated task, can leave broad and sometimes unexpected traces in a model's representations and behavior (25). Directional-overlap and ceiling-robust metrics indicate that the fingerprint and the honesty-related direction are largely, but not completely, dissociable, and the residual entanglement (log-loss increases of 3.68 percent to 23.38 percent after removing the fingerprint direction) means the confound cannot be fully ruled out on the basis of cosine similarity alone. Any study that constructs contrast pairs by fine-tuning two model variants, rather than by contrasting prompts or naturally occurring completions, should treat this control as a required step rather than an optional robustness check.

The paraphrase robustness results temper the positive separability findings discriminability. The reduction in AUC from near ceiling to 0.52 to 0.59 under lexical, structural, and back-translation paraphrasing suggests that the linear direction identified here is substantially more dependent on the surface form of the training questions than on a more abstract representation of the honest/dishonest distinction. This pattern is consistent with a growing body of evidence that linear probes for deception- and honesty-related properties generalize unevenly outside their training distribution. Probes trained on simple contrastive honesty prompts have been shown to transfer with high fidelity to some out-of-distribution deception settings (11) but to fail, sometimes reaching near-chance balanced accuracy, on other categories of deceptive or dishonest behavior that differ from the training distribution in kind rather than in surface form (26). The present results extend this concern to a setting where the distribution shift is much narrower, a reworded version of the same question, which suggests that fragility to distribution shift in this literature is not limited to qualitatively different deception types but can appear even under superficial input variation.

The activation-steering results do not provide evidence for or against a causal role of the identified direction in production. Norm-calibrating the intervention magnitude removed the exact-zero effect observed in the earlier, non-calibrated pilot, but the resulting effects were small relative to sampling noise, non-monotonic in the intervention magnitude, and accompanied by visible degradation in output fluency at the magnitudes required to produce any measurable shift. This trade-off between intervention strength and output coherence has been documented in the broader activation-steering literature (21) and is a recognized limitation of naive additive steering rather than a finding specific to this study. The absence of a clean causal effect is not interpreted as evidence that the identified direction is causally inert; a more targeted intervention protocol, for example, steering at a narrower range of layers, using smaller step sizes between the tested magnitudes, or filtering generations for fluency before scoring, may be more diagnostic, and sufficient resources were not available to conduct that analysis within the present study.

Taken together, these results have a specific and limited implication for the broader question of deceptive alignment. They show that a behavioral distinction directly produced through supervised fine-tuning is easy to detect linearly, at least in most of the architectures tested, and that this detectability survives moderate distribution shift (TruthfulQA to MMLU) but not surface-level paraphrasing of the same questions. This is a different and, it is believed, easier detection problem than identifying strategic deception in a model that was never directly trained to produce the deceptive output, or that maintains a competing accurate representation while producing an inaccurate one. Prior work on backdoor-induced deception (10) and on strategic deception in more naturalistic multi-turn settings (11) addresses closer analogues of that harder problem, and the present results should not be read as evidence about detection performance in those settings.

**Limitations**

- The adapters used in the causal-intervention (Section 3.12/4.10) and paraphrase-robustness (Section 3.13/4.9) experiments were fine-tuned with a LoRA configuration that may differ from the main fine-tuning pipeline due to loss of the original compute environment. This should be explicitly noted as limiting the interpretation of these two supplementary experiments; it does not affect the main probe results in Sections 4.1–4.2.
- The LoRA fingerprint identified in Section 4.6 is a genuine finding (AUC ≈ 0.99–1.00), not an artifact, and constrains how strongly the main honesty/dishonesty separability results can be attributed to response content alone; see Section 4.6 for the directional-overlap analysis used to partially address this.
- The paraphrase-robustness results (Section 4.9) indicate the identified direction is substantially tied to surface-level question phrasing and does not transfer robustly to reworded but semantically equivalent questions.
- A norm-calibrated activation-steering experiment (Section 4.10) resolved the exact-zero effect observed in an earlier, non-calibrated pilot, but the resulting effects remain small (Δ incorrect-rate 0.017–0.250), non-monotonic in $\alpha$, and were accompanied, in the manually inspected example generations, by an apparent loss of fluency at the magnitudes required to produce a measurable change. The causal role of the identified linear direction in the model's output generation therefore remains unresolved.
- Model scale is limited to 1.4–9B parameters; whether these findings extend to substantially larger models remains an open question.

**Conclusion**

Supervised fine-tuning of incorrect responses induces activation representations that, in most architectures tested, are linearly separable from their honest counterparts in the first few layers and are transferred with high fidelity to a held-out domain. This separability, by itself, is not sufficient evidence that the underlying signal reflects the content of the honest/dishonest distinction: a fine-tuning process applied to unrelated data produces a relatively strong linear signature, the identified direction does not transfer to paraphrased versions of the same questions, and a norm-calibrated steering experiment does not establish a reliable causal role for that direction in generation. Positive and negative results are reported together because both are considered necessary to describe what supervised fine-tuning of incorrect responses leaves in the activation space and what it does not. The main practical recommendation of this work is methodological: studies that use fine-tuned model variants to construct behavioral contrast pairs for probing should include a fine-tuning fingerprint control and, where feasible, a distribution-shift check beyond a single held-out domain, since near-ceiling in-distribution AUC alone is not a reliable indicator that a probe has recovered a semantically meaningful direction.

**Data and Code Availability**
All probe training scripts and evaluation code are publicly available at https://github.com/vzm1399/synthetic-dishonesty-probing.

**Competing Interests**
None.

**Acknowledgments**

This work was conducted as part of the Algoverse AI Safety Research Program. The author thanks Sunishchal Dev for helpful comments and feedback. ChatGPT was used for language editing and grammar checking.

## References

1. Hubinger E, Van Merwijk C, Mikulik V, Skalse J, Garrabrant S. Risks from learned optimization in advanced machine learning systems. arXiv preprint arXiv:190601820. 2019.
2. Hubinger E, Denison C, Mu J, Lambert M, Tong M, MacDiarmid M, et al. Sleeper agents: Training deceptive llms that persist through safety training. arXiv preprint arXiv:240105566. 2024.
3. Greenblatt R, Denison C, Wright B, Roger F, MacDiarmid M, Marks S, et al. Alignment faking in large language models. arXiv preprint arXiv:241214093. 2024;2(1).
4. Pacchiardi L, Chan A, Mindermann S, Moscovitz I, Pan A, Gal Y, et al., editors. How to catch an ai liar: Lie detection in black-box llms by asking unrelated questions. International Conference on Learning Representations; 2024.
5. Burns C, Ye H, Klein D, Steinhardt J. Discovering latent knowledge in language models without supervision. arXiv preprint arXiv:221203827. 2022.
6. Marks S, Tegmark M. The geometry of truth: Emergent linear structure in large language model representations of true/false datasets. arXiv preprint arXiv:231006824. 2023.
7. Park K, Choe YJ, Veitch V. The linear representation hypothesis and the geometry of large language models. arXiv preprint arXiv:231103658. 2023.
8. Hu EJ, Shen Y, Wallis P, Allen-Zhu Z, Li Y, Wang S, et al. Lora: Low-rank adaptation of large language models. arXiv preprint arXiv:210609685. 2021.

9. Zou A, Phan L, Chen S, Campbell J, Guo P, Ren R, et al. Representation engineering: A top-down approach to ai transparency. arXiv preprint arXiv:231001405. 2023.
10. MacDiarmid M, Maxwell T, Schiefer N, Mu J, Kaplan J, Duvenaud D, et al. Simple probes can catch sleeper agents. Anthropic Research Updates. 2024.
11. Goldowsky-Dill N, Chughtai B, Heimersheim S, Hobbhahn M, editors. Detecting strategic deception with linear probes. Forty-second International Conference on Machine Learning; 2025.
12. Meng K, Bau D, Andonian A, Belinkov Y. Locating and editing factual associations in gpt. Advances in neural information processing systems. 2022;35:17359-72.
13. Mitchell E, Lin C, Bosselut A, Finn C, Manning CD. Fast model editing at scale. arXiv preprint arXiv:211011309. 2021.
14. Goodfellow IJ, Shlens J, Szegedy C. Explaining and harnessing adversarial examples. arXiv preprint arXiv:14126572. 2014.
15. Madry A, Makelov A, Schmidt L, Tsipras D, Vladu A. Towards deep learning models resistant to adversarial attacks. arXiv preprint arXiv:170606083. 2017.
16. Lin S, Hilton J, Evans O, editors. Truthfulqa: Measuring how models mimic human falsehoods. Proceedings of the 60th annual meeting of the association for computational linguistics (volume 1: long papers); 2022.
17. Hendrycks D, Burns C, Basart S, Zou A, Mazeika M, Song D, et al. Measuring massive multitask language understanding. arXiv preprint arXiv:200903300. 2020.
18. Dettmers T, Pagnoni A, Holtzman A, Zettlemoyer L. Qlora: Efficient finetuning of quantized llms. Advances in neural information processing systems. 2023;36:10088-115.
19. Dao T, editor Flashattention-2: Faster attention with better parallelism and work partitioning. International Conference on Learning Representations; 2024.
20. Guo C, Pleiss G, Sun Y, Weinberger KQ, editors. On calibration of modern neural networks. International conference on machine learning; 2017: PMLR.
21. Rimsky N, Gabrieli N, Schulz J, Tong M, Hubinger E, Turner A, editors. Steering llama 2 via contrastive activation addition. Proceedings of the 62nd Annual Meeting of the Association for Computational Linguistics (Volume 1: Long Papers); 2024.
22. Pedregosa F, Varoquaux G, Gramfort A, Michel V, Thirion B, Grisel O, et al. Scikit-learn: Machine learning in Python. the Journal of machine Learning research. 2011;12:2825-30.
23. Wolf T, Debut L, Sanh V, Chaumond J, Delangue C, Moi A, et al., editors. Transformers: State-of-the-art natural language processing. Proceedings of the 2020 conference on empirical methods in natural language processing: system demonstrations; 2020.
24. Engels J, Michaud E, Liao I, Gurnee W, Tegmark M, editors. Not all language model features are one-dimensionally linear. International Conference on Learning Representations; 2025.
25. Betley J, Tan D, Warncke N, Sztyber-Betley A, Bao X, Soto M, et al. Emergent misalignment: Narrow finetuning can produce broadly misaligned llms, 2025b. URL https://arxiv org/abs/250217424. 2025:8.
26. Kretschmar K, Laurito W, Maiya S, Marks S. Liars' Bench: Evaluating Lie Detectors for Language Models. arXiv preprint arXiv:251116035. 2025.

## Appendix A — Data Leakage Diagnostic (LoRA Fingerprint Control)

During the initial LoRA fingerprint control described in Section 3.11, cross-validation using standard StratifiedKFold produced fingerprint-probe AUC values that were suspiciously close to a ceiling of 1.0, even in control conditions constructed to be uninformative (two models fine-tuned on random, honesty-unrelated data splits). Investigation traced this to near-duplicate or exactly duplicate pooled activation vectors, which can arise when the mean-pooling procedure is applied to short or templated responses. When such duplicate or near-duplicate vectors were split across the training and held-out folds under StratifiedKFold, a probe could achieve high held-out accuracy by matching a held-out vector to an essentially identical training vector rather than by learning a generalizable direction. This represents a data-leakage pathway rather than genuine separability.

To confirm this mechanism directly, activation vectors were compared pairwise for exact (bitwise) equality across the dataset, and pairs found to be bitwise-identical were cross-referenced against the corresponding StratifiedKFold fold assignment to determine whether members of the same duplicate cluster had been split across training and test folds within a given cross-validation run.

Replacing StratifiedKFold with GroupKFold, grouped by a duplicate-cluster identifier so that all members of a near-duplicate cluster are guaranteed to fall in the same fold, removed this leakage pathway. All fingerprint-control and honesty-probe results reported in this manuscript use the GroupKFold-corrected procedure.